\documentclass[lettersize,journal]{IEEEtran}
\usepackage{amsmath,amsfonts}
\usepackage{algorithmic}
\usepackage{algorithm}
\usepackage{array}
\usepackage[caption=false,font=normalsize,labelfont=sf,textfont=sf]{subfig}
\usepackage{textcomp}
\usepackage{stfloats}
\usepackage{url}
\usepackage{verbatim}
\usepackage{graphicx}
\usepackage{cite}
\usepackage{adjustbox}
\usepackage[colorlinks,
            linkcolor=blue,
            anchorcolor=blue,
            citecolor=blue, 
            urlcolor=blue]{hyperref}
\usepackage{orcidlink}

\begin{document}

\title{StructGS: Adaptive Spherical Harmonics and Rendering Enhancements for Superior 3D Gaussian Splatting}

\author{Zexu Huang\orcidlink{0009-0006-6251-9694}, Min Xu\orcidlink{0000-0001-9581-8849},~\IEEEmembership{Member,~IEEE}, Stuart Perry\orcidlink{0000-0002-2794-3178},~\IEEEmembership{Senior Member,~IEEE}
\thanks{Zexu Huang, Min Xu and Stuart Perry are with the Perceptual Imaging Laboratory (PILab), School of Electrical and Data Engineering, University of Technology Sydney, Ultimo, NSW 2007, Australia (email: \href{mailto:zexu.huang@student.uts.edu.au}{zexu.huang@student.uts.edu.au}; \href{mailto:min.xu@uts.edu.au}{min.xu@uts.edu.au}; \href{mailto:stuart.perry@uts.edu.au}{stuart.perry@uts.edu.au})}
\thanks{Manuscript received Dec 3, 2024}}



\maketitle

\begin{abstract}
Recent advancements in 3D reconstruction coupled with neural rendering techniques have greatly improved the creation of photo-realistic 3D scenes, influencing both academic research and industry applications. The technique of 3D Gaussian Splatting and its variants incorporate the strengths of both primitive-based and volumetric representations, achieving superior rendering quality. While 3D Geometric Scattering (3DGS) and its variants have advanced the field of 3D representation, they fall short in capturing the stochastic properties of non-local structural information during the training process. Additionally, the initialisation of spherical functions in 3DGS-based methods often fails to engage higher-order terms in early training rounds, leading to unnecessary computational overhead as training progresses. Furthermore, current 3DGS-based approaches require training on higher resolution images to render higher resolution outputs, significantly increasing memory demands and prolonging training durations. We introduce StructGS, a framework that enhances 3D Gaussian Splatting (3DGS) for improved novel-view synthesis in 3D reconstruction. StructGS innovatively incorporates a patch-based SSIM loss, dynamic spherical harmonics initialisation and a Multi-scale Residual Network (MSRN) to address the above-mentioned limitations, respectively. Our framework significantly reduces computational redundancy, enhances detail capture and supports high-resolution rendering from low-resolution inputs. Experimentally, StructGS demonstrates superior performance over state-of-the-art (SOTA) models, achieving higher quality and more detailed renderings with fewer artifacts.  (The link to the code will be made available after publication)

\end{abstract}

\section{Introduction}
\IEEEPARstart{R}{ecent} advancements in 3D reconstruction have enhanced novel-view synthesis, allowing for the creation of photorealistic representations of volumetric scenes. Neural Radiance Fields (NeRF)~\cite{mildenhall2021nerf} represents a significant milestone in 3D rendering which employs a multi-layer perceptron (MLP) to produce high-quality, geometrically coherent images from novel viewpoints. However, this method incurs the drawback of time-consuming stochastic sampling, which can lead to slower performance and potential noise artefacts. Its variants primarily enhance rendering quality~\cite{barron2021mip, barron2022mip} and accelerate the convergence speed~\cite{muller2022instant, sun2022direct}.

Due to limitations in rendering quality and training speeds of NeRF-based models, recent breakthroughs in 3D Gaussian Splatting (3DGS)~\cite{kerbl20233d} have significantly improved these aspects. This method fine-tunes a sequence of 3D Gaussians, initially shaped using Structure-from-Motion (SfM)~\cite{schonberger2016structure} to effectively model scenes with volumetric continuity. This enables faster rasterization through projection onto 2D planes. Nonetheless, when camera angles deviate from the original training configurations or during close-up views, 3DGS often produces visual distortions due to inadequate detail resolution. To overcome these limitations, recent 3DGS variants~\cite{yu2024mip, lu2024scaffold} have introduced a 3D smoothing filter to control extreme frequency variations and employ anchor points to set up 3D Gaussians. These enhancements significantly improve visual accuracy and versatility in a range of applications. Despite significant advancements in 3DGS, current models typically assess training performance by simply comparing rendered images against ground truths using individual-pixel Structural Similarity (SSIM). However, it lacks a comprehensive assessment of structural similarities, potentially overlooking subtle yet critical topological discrepancies. Additionally, the initialisation of spherical harmonics in 3DGS-based models typically involves setting the zeroth-degree coefficient (dc term) to the RGB colour of the ground truth while all remaining terms are initialised to zero. As shown in Tab.~\ref{tab:3dgs_variance}, such a method may lead to higher dimensions uninitialised in the early iterations and calculation redundancies and biases in higher-order spherical harmonics which compromises the model's efficiency. Furthermore, current 3DGS-based methods that render higher resolution images necessitate training with higher resolution images, which not only increases memory demands but also significantly increases the computation complexity.

\begin{figure*}[!t]
    \centering
    \includegraphics[width=\textwidth]{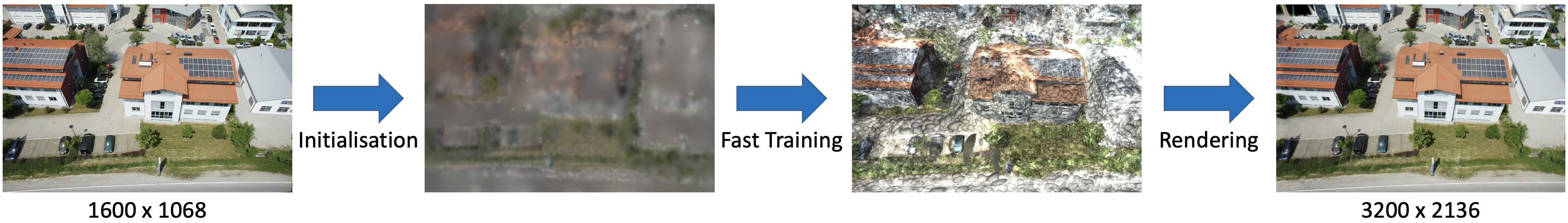}
    \caption{StructGS: We present a framework that produces high-quality renderings from a set of single-view camera images. Following scene reconstruction, our approach allows for rapid rendering at resolutions higher than that of the input image. To achieve this, we leveraged a patch SSIM loss and a total variation (TV) loss regularizer, which effectively capture nonlocal structural information and enhance image smoothness. Additionally, we proposed a dynamic adjustment strategy for spherical harmonics based on the opacity and distance of Gaussian spheres. We also integrated a pre-trained Multi-scale Residual Network to facilitate super-resolution rendering..}
    \label{fig:intro}
\end{figure*}

To address these challenges, we introduce StructGS, a framework designed to enhance rendering quality, reduce redundancy and bias in higher-order spherical harmonics and achieve higher-resolution images. Following the NeRF method~\cite{xie2023s3im}, StructGS uses a patch SSIM loss to capture nonlocal structural similarity from stochastically sampled pixels, combined with a TV loss regulariser for smoothness. To refine spherical harmonics initialisation and optimisation, we design a dynamic adjustment strategy that considers the transparency of Gaussian spheres. The first three RGB dimensions of the spherical harmonics are initialised and optimised based on the transparency levels of each sphere. Additionally, the remaining harmonics are adjusted dynamically using distance information: higher-order harmonics are used for points further from neighbours to capture finer details, while closer points use lower-order harmonics for broader features. During the rendering process, StructGS incorporates a pre-trained Multi-scale Residual Network~\cite{li2018multi} to perform super-resolution on the rendered images, which enables the model to train on low-resolution input images and then produce high-resolution, high-quality rendered images.

Through rigorous experiments, StructGS has demonstrated superior performance over current state-of-the-art (SOTA) 3DGS-based and previous 3D reconstruction models. It effectively reduces the calculation redundancy of higher-order spherical harmonics, achieving better quality with fewer training iterations. Moreover, the designed framework enhanced by loss functions and a pre-trained Multi-Scale Residual Network (MSRN)~\cite{li2018multi} enables StructGS to render enhanced quality and higher resolution images. (see Fig.~\ref{fig:intro})

\subsection*{Contributions}
In summary, the contributions of this work are as follows:

\begin{enumerate}
    \item Different from existing methods only considering individual pixel-based structural similarity, we leveraged a patch SSIM loss and a TV loss regularizer to effectively capture nonlocal structural information and enhance image smoothness. This approach allows for a more refined representation of structural details in 3D reconstructions.

    \item We proposed a dynamic adjustment strategy for spherical harmonics based on the opacity and distance of Gaussian spheres from each other. This method optimises the initialisation and adjustment of spherical harmonics, reducing redundancy and bias in higher-order components, and improving the quality of training within fewer iterations.

    \item We incorporated a pre-trained Multi-scale Residual Network for super-resolution, StructGS is capable of producing high-resolution, high-quality images from low-resolution inputs. This advancement enhances the framework's ability to handle detailed textures and complex geometries.

\end{enumerate}

The organization of this paper is as follows: Section~\ref{sec:literatures} provides a review of prior research pertinent to our study. Section~\ref{sec:method} describes the methods utilized in our research. Section~\ref{sec:experiment} details our experimental setup, describes our model implementations, compares the performance of our model with other advanced 3D reconstruction methods and discusses the results of ablation studies as well as the limitations of our approach. The paper is concluded in Section~\ref{sec:conclusion}, where we summarize the principal contributions of our work.
\begin{table}
\begin{center}
\caption{Variance of spherical harmonics terms of different degrees for two scenes (bicycle, bonsai) from Mip-NeRF 360 dataset~\cite{barron2022mip} and one villa scene at various training iterations of the original 3D Gaussian Splatting~\cite{kerbl20233d}. The results show that many higher-degree spherical harmonics terms are not initialised at early training iterations. Addressing this issue can improve the model's performance during the early stages of training.}
\label{tab:3dgs_variance}
\begin{adjustbox}{width=3.45in,center}
\begin{tabular}{lcccc}
\hline
\multicolumn{5}{c}{\textbf{Variance of Spherical Harmonics Terms}} \\ 
Scene & degree 0 & degree 1 & degree 2 & degree 3 \\
\hline
bicycle (1k iteration) & 1.321 & 0.0 & 0.0 & 0.0 \\
bicycle (2k iteration) & 1.624 & 1.472$\times10^{-4}$ & 0.0 & 0.0 \\
bicycle (3k iteration) & 1.501 & 3.592$\times10^{-4}$ & 9.451$\times10^{-5}$ & 0.0 \\
\hline
bonsai (1k iteration) & 0.712 & 0.0 & 0.0 & 0.0 \\
bonsai (2k iteration) & 0.860 & 1.217$\times10^{-4}$ & 0.0 & 0.0 \\
bonsai (3k iteration) & 0.894 & 2.998$\times10^{-4}$ & 1.020$\times10^{-4}$ & 0.0 \\
\hline
villa (1k iteration) & 1.442 & 0.0 & 0.0 & 0.0 \\
villa (2k iteration) & 1.477 & 4.895$\times10^{-4}$ & 0.0 & 0.0 \\
villa (3k iteration) & 1.440 & 0.001 & 3.958$\times10^{-4}$ & 0.0 \\
\hline
\end{tabular}
\end{adjustbox}
\end{center}
\end{table}

\section{Related Work}
\label{sec:literatures}
\noindent \textbf{Novel View Synthesis}\ \ \ \
Image-based rendering methodologies have traditionally been employed in the field of novel view synthesis for real-world scenes. Techniques such as Structure-from-Motion (SfM)~\cite{schonberger2016structure} play a pivotal role by facilitating the estimation of camera parameters using a series of images. These parameters are crucial for accurately projecting the colours of input images onto a new viewpoint~\cite{debevec1998efficient}. This method fundamentally depends on the construction of approximate geometry which typically involves creating mesh models or point clouds. To enhance these models, refinements are often introduced by employing Multi-View Stereo (MVS) processes~\cite{goesele2007multi, schonberger2016pixelwise}. Although these methods are effective, real-world data frequently exhibit issues such as inaccuracies in camera calibration and geometric errors~\cite{shum2000review}. These problems result in undesirable outcomes during the rendering process, including artifacts at object borders and diminished sharpness of details. To address these challenges, recent advancements have incorporated neural rendering techniques~\cite{tewari2022advances}, which significantly mitigate such artifacts and improve the fidelity of the synthesized views. Some approaches use rendering techniques like viewpoint interpolation~\cite{qin2022bullet} or depth sensors to capture scene geometry directly~\cite{alexiadis2012real}. However, these methods are typically limited to indoor environments. This restriction is due to the sensor's range constraints and their primary focus on adapting 2D content.

\vspace{1em}

\noindent \textbf{MLP-based Radiance Fields}\ \ \ \
Initial approaches in neural fields have employed multi-layer perceptrons (MLPs) as the primary mechanism for approximating the geometry and visual characteristics of 3D scenes. In these systems, spatial coordinates and viewing directions are fed into the MLP. The MLP then computes attributes such as the signed distance from the scene surface (SDF)~\cite{park2019deepsdf, wang2021neus, wang2023neus2} or determines the colour and density at a specific coordinate~\cite{mildenhall2021nerf, barron2021mip, barron2022mip, huang2023natural}. While MLP-based methods leverage their volumetric capabilities to achieve state-of-the-art results in novel view synthesis, they face significant challenges. The primary issue is the necessity for the MLP to process a vast number of point samples along each camera ray, which drastically slows down rendering and limits the ability to handle large and complex scenes effectively. Recent research efforts have aimed to enhance the efficiency and broaden the scalability of Neural Radiance Fields (NeRF). These advancements~\cite{muller2022instant, chen2022tensorf, sun2022direct, huang2024efficient, fridovich2022plenoxels} have involved the adoption of discrete or sparse volumetric structures, such as voxel grids and hash tables. These structures are essential because they incorporate learnable features that function as positional encodings for three-dimensional coordinates. Moreover, these approaches~\cite{barron2022mip, chen2022tensorf, yu2021plenoctrees} apply hierarchical sampling strategies and they utilize techniques for low-rank approximations. However, the ongoing dependence on volumetric ray marching poses compatibility issues with standard graphics platforms, which are traditionally optimized for polygonal rendering.

\vspace{1em}

\noindent \textbf{Point-based Radiance Fields}\ \ \ \
In point-based radiance field rendering, using point clouds as explicit proxy representations offers significant advantages. These point clouds are efficiently captured using technologies like LiDAR~\cite{liao2022kitti} and Structure-from-Motion/Multi-View Stereo methods~\cite{schonberger2016pixelwise}. Comprising an unstructured assortment of spatial samples with varied neighbour distances, point clouds faithfully represent the original data captured. The process of rendering these point clouds is generally rapid~\cite{schutz2019real}. Enhancements using neural descriptors~\cite{ruckert2022adop} or specially optimized attributes~\cite{kopanas2021point} allow for high-quality visual outputs through differentiable point renderers~\cite{wiles2020synsin, arvanitis2021broad}. Nevertheless, the discrete rasterization process can lead to visual artifacts like aliasing or overdrawing, particularly when multiple points converge on the same pixel. Recently, the 3DGS framework~\cite{kerbl20233d, jiang2024gs, yu2024mip, lu2024scaffold} innovatively applies directionally dependent 3D Gaussians for rendering three-dimensional scenes. This approach leverages Structure from Motion (SfM)~\cite{schonberger2016structure} to set up the initial 3D Gaussian structures which are then refined and optimized as volumetric models. This approach combines anisotropic Gaussians with a high-speed tiled renderer, optimizing the sizes of splats through gradient descent. However, it is crucial to limit the number of Gaussians to prevent performance degradation. This constraint may result in the over-blurring of small, detailed elements. 

\begin{figure*}[!t]
    \centering
    \includegraphics[width=1\textwidth]{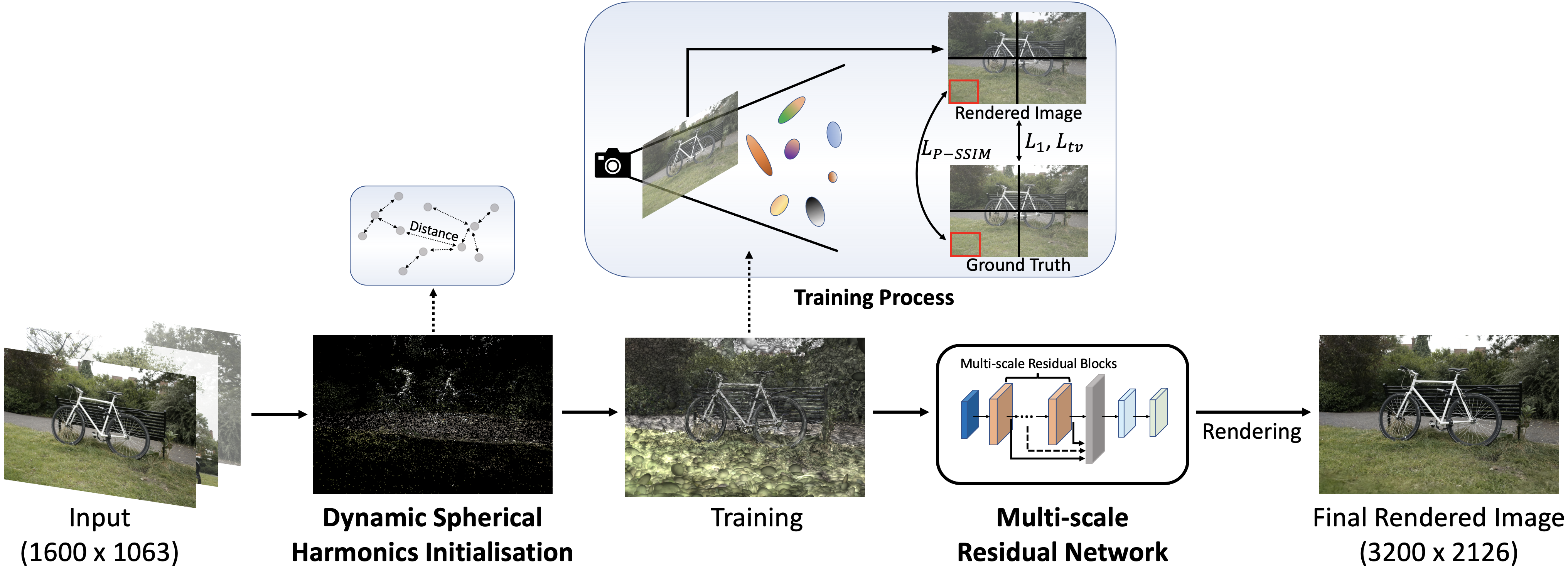}
    \caption{\textbf{Overview of StructGS:} In the initialisation phase, our model employs a \textbf{dynamic adjustment of spherical harmonics} based on opacity weighting to optimise the first three RGB dimensions for each Gaussian sphere. Distance information further refines initialisation, with higher-order harmonics capturing more details for distant points and lower-order for nearer points. During the training phase, the rendered and ground truth images are divided into several patches. Within these patches, the SSIM loss ($L_{SSIM}$) for small areas is calculated using a kernel. The results are then summed and averaged. The rendered and ground truth images are also assessed for total variation loss ($L_{tv}$) and L1 loss ($L_{1}$). After training, our model incorporates a pre-trained \textbf{Multi-scale Residual Network} to render high-quality and high-resolution images.}
    \label{fig:model_structure}
\end{figure*}

Recent 3DGS variants~\cite{yu2024mip, lu2024scaffold} have implemented a 3D smoothing filter to manage abrupt frequency changes and utilize anchor points for establishing 3D Gaussians. They~\cite{yu2024mip, lu2024scaffold} successfully eliminated some artifacts at high frequencies in 3DGS and also enhanced the rendering quality. Despite significant progress in 3DGS, current models~\cite{kerbl20233d, yu2024mip, lu2024scaffold} often evaluate training performance by simply comparing rendered images with ground truths using SSIM at the pixel level, potentially missing subtle yet crucial topological differences and often omitting important structural information. Moreover, initializing spherical harmonics in 3DGS models typically sets the DC term to the ground truth's RGB color, with all other terms set to zero, leading to redundancies and biases in higher-order harmonics that reduce model efficiency. Furthermore, 3DGS methods rendering higher resolution images require training on higher resolution images, which increases memory demands and significantly prolongs training time. As shown in Fig.~\ref{fig:model_structure}, our method addresses these issues by incorporating non-local structural information and a total variation (TV) regulariser term during the loss calculation. Additionally, we dynamically adjust the initialisation of spherical harmonics to reduce computational redundancies. Furthermore, we integrate a pre-trained Multi-scale Residual Network to enhance the rendering quality and resolution, thus avoiding the increased training time and computational costs previously required for training on high-resolution images.
\section{Methodology}
\label{sec:method}
\subsection{Preliminary}
Previous research~\cite{kerbl20233d} introduced a method to represent 3D scenes using a group of scaled 3D Gaussian $\{G_n | n = 1, \dots, N\}$ and render images via splatting. Each Gaussian primitive $G_n$ is characterized by an opacity $\alpha_n \in [0, 1]$, a center position $\mathbf{q}_n \in \mathbb{R}^{3 \times 1}$, and a covariance matrix $\Sigma_n \in \mathbb{R}^{3 \times 3}$ defined in world space. The form of each scaled Gaussian function is given by:
\begin{equation}
G_n(\mathbf{x}) = e^{-\frac{1}{2} (\mathbf{x} - \mathbf{q}_n)^T \Sigma_n^{-1} (\mathbf{x} - \mathbf{q}_n)}.
\end{equation}
To ensure that $\Sigma_n$ remains a valid covariance matrix, it is expressed using a positive semi-definite parameterization: $\Sigma_n = \mathbf{U}_n \mathbf{d}_n \mathbf{d}_n^T \mathbf{U}_n^T$. In this formulation, $\mathbf{d} \in \mathbb{R}^3$ represents a scaling vector while $\mathbf{U} \in \mathbb{R}^{3 \times 3}$ is a rotation matrix defined by a quaternion for orientation~\cite{kerbl20233d}. Rendering an image from a given viewpoint requires transforming these 3D Gaussians $\{G_n\}$ into camera space, defined by a rotation matrix $\mathbf{C} \in \mathbb{R}^{3 \times 3}$ and a translation vector $\mathbf{t} \in \mathbb{R}^3$. The centre and covariance matrix are transformed as follows:
\begin{equation}
\mathbf{q}_n' = \mathbf{C} \mathbf{q}_n + \mathbf{t}, \quad \Sigma_n' = \mathbf{C} \Sigma_n \mathbf{C}^T.
\end{equation}
Then, a local transformation projects these transformed Gaussians into ray space using a Jacobian matrix $J_n$, which approximates the projective transformation at the Gaussian centre $\mathbf{q}_n'$. This transformation results in an updated covariance matrix:
\begin{equation}
\Sigma_n'' = J_n \Sigma_n' J_n^T.
\end{equation}
By eliminating the third row and column of $\Sigma_n''$, we obtain a 2D covariance matrix $\Sigma_n^{2D}$, which enables the efficient modelling of 2D Gaussians, denoted as $G_n^{2D}$. Eventually, 3D Gaussian Splatting (3DGS)~\cite{kerbl20233d} uses spherical harmonics to model view-dependent colour $c_n$, and renders the image through alpha blending based on the Gaussian's depth order (1 to $N$):
\begin{equation}
c(\mathbf{x}) = \sum_{n=1}^N c_n \alpha_n G_n^{2D}(\mathbf{x}) \prod_{j=1}^{n-1} (1 - \alpha_j G_j^{2D}(\mathbf{x})),
\end{equation}
where:
\begin{equation}
    G_n^{2D}(\mathbf{x}) = e^{-\frac{1}{2} (\mathbf{x} - \mathbf{q}_n)^T (\Sigma_n^{2D} + h \mathbf{I})^{-1} (\mathbf{x} - \mathbf{q}_n)}.
\end{equation}
The symbol $\mathbf{I}$ denotes a 2D identity matrix and $h$ serves as a scalar hyperparameter for dilation.
\subsection{Non-Local Structural Information in SSIM}
Current 3DGS-based methods~\cite{kerbl20233d, yu2024mip, lu2024scaffold} generally measure training performance by contrasting rendered images with ground truths through pixel-level Structural Similarity (SSIM). However, this method falls short of capturing all structural nuances, potentially missing key topological variances. As shown in Fig.~\ref{fig:model_structure}, our model undergoes a specific iteration procedure during the training phase to compute the Structural Similarity Index (SSIM) over stochastic patch pairs derived from both the rendered image $R_i$ and the Ground Truth image $G_i$. Each image is segmented into $p \times p$ patches after $k$ iterations. For each patch, a rendered patch is formed from a random selection of pixels within the patch and SSIM is computed for these stochastic patch pairs using a $K \times K$ kernel. The stride size for this operation is denoted as $s$. The SSIM is computed for each patch pair, and the process continues until partial SSIM values are calculated for the entire image. The summation of partial SSIM values is then averaged to obtain the final Patch-based SSIM (P-SSIM). Mathematically, the P-SSIM can be expressed as:
\begin{equation}
\text{P-SSIM}(r_{i}, g_{i}) = \frac{1}{P} \sum_{i=1}^{P} \text{SSIM}(r_{i}, g_{i}),
\end{equation}
where $P$ represents the number of stochastic patch pairs assessed, and $r_{i}$ and $g_{i}$ represent the individual patches of the rendered and ground truth images, respectively. Given that SSIM values range between $[-1, 1]$, we define our loss based on P-SSIM as:
\begin{equation}
L_{\text{P-SSIM}} = 1 - \text{P-SSIM}(r_{i}, g_{i}) = 1 - \frac{1}{P} \sum_{i=1}^{P} \text{SSIM}(r_{i}, g_{i}).
\end{equation}
Following the previous work~\cite{xie2023s3im}, we set $K = 4$ and the stride $S = K$ without further fine-tuning. The patch size $P$ is experimentally determined to be 10. Through the experiment, we set $k$ = 25000 in our experiment.

While the original SSIM is differentiable and allows for gradient-based optimisation, direct optimisation of SSIM can lead to suboptimal results. Unlike previous 3DGS~\cite{kerbl20233d} approaches that used SSIM directly with local patches, our method involves stochastic patch pairs that capture non-local structural information across the image which can produce superior results. This is demonstrated through our ablation studies which confirm that optimising P-SSIM via stochastic patches significantly outperforms conventional local patch-based SSIM optimisation. This approach captures relationships between distant pixels, thus providing a method for enhancing the quality of rendered images.
\subsection{Dynamic Adjustment of Spherical Harmonics Initialisation}
The conventional initialisation of spherical harmonics (SH) in 3D Gaussian Splatting typically sets the zeroth-degree coefficient (dc term) of the SH to the initial RGB values of the point cloud which leaves higher dimensions uninitialised in the early iterations. This approach often results in unnecessary computational overhead to optimise higher-order spherical harmonics dimensions. To address these inefficiencies and biases, we propose a method for dynamically adjusting the initialisation of spherical harmonics based on the opacity and distance metrics of the points within the point cloud data. Initially, the spherical harmonics coefficients are calculated based on the RGB colour data of the point cloud.  The first three SH coefficients are initialised using the RGB values weighted by the opacity $\alpha_n$ of each point which allows these primary coefficients to encapsulate the fundamental colour information modified by the transparency of each point:
\begin{equation}
    \text{SH}_{\text{RGB}}(0) = \text{RGB} \times \alpha_n.
\end{equation}
Here, $\alpha_n$ is the opacity derived from the inverse sigmoid function. This design modifies the transparency based on the point's attributes; the more transparent the point, the less significant the RGB values become, thereby tailoring the primary colour information embedded in the spherical harmonics.

The distance metric from each point to its neighbours significantly influences the level of detail needed in the spherical harmonics representation. Points that are closer to their neighbours are initialised with lower-degree SH terms, as the fine details are less discernible at small distances. Conversely, points that are farther away are initialised with higher-degree SH terms to capture more detail. Gaussians with close neighbours can rely on neighbours for fine colour variations to be represented, whereas Gaussians with large distances from their neighbours need more SH to represent variations in light with angle. The coefficients $\nu$ of the SH can be expressed as:
\begin{equation}
     \text{D} = \max(1, \min(d, M)), \quad \nu = (\text{D} + 1)^2,
\end{equation}
where $d$ is the scaled Euclidean distance between points and $M$ is a parameter that represents the upper bond degree of the spherical harmonics. Through our experiment, we set $M$ as 5. The degree $D$ of spherical harmonics is dynamically set to ensure detailed capture for distant points and less complexity for closer points. The spherical harmonics are dynamically adjusted based on both the opacity and the calculated distance. The method combines these attributes to initialise the non-zero coefficients of SH for higher-order terms. For each point, its corresponding spherical harmonics coefficients beyond the first are initialised as:
\begin{equation}
    \text{SH}_{\text{RGB}}(\nu) = \text{RGB} \times \alpha_n \times (1 - e^{-s d}),
\end{equation}
where $n$ is the index for spherical harmonics coefficients, $s$ is a scaling factor adjusting the impact of distance on higher-order terms, and $e^{-s d}$ modulates the influence of distance on these terms. This dynamic initialisation approach seeks to optimise computational resources and accelerate the training process. Results demonstrate enhanced performance with fewer iterations required in training.
\subsection{Multi-scale Residual Network}
Current methods based on 3DGS-based methods~\cite{kerbl20233d, yu2024mip, lu2024scaffold} that render higher resolution images require training with higher resolution images. This not only increases memory usage but also significantly prolongs the training time. Our model addressed the limitation by employing a pre-trained Multi-scale Residual Network (MSRN)~\cite{li2018multi} $F_\theta$ to fit into the rendering process in 3DGS. It processes low-resolution (LR) pixel represented as \(p^{(\text{LR})}\) to generate super-resolution (SR) pixel denoted as \(p^{(\text{SR})}\). The input image undergoes a feature extraction process where the MSRN utilises $K$ multi-scale residual blocks (MSRBs) to capture intrinsic patterns at various scales and depths. The LR input is fed directly into the network, avoiding initial upsampling, thus preserving the original low-resolution data for the learning process.

\noindent\textbf{Multi-scale Residual Block (MSRB)}\ \ \ \
To effectively identify features across multiple scales, the MSRB integrates two critical components: multi-scale feature fusion and local residual learning. In the process of multi-scale feature fusion, it leverages a dual-pathway architecture. Each pathway utilises convolutional kernels of different sizes, allowing for the interplay and fusion of feature maps at varying granularities. During the local residual learning procedure, each MSRB facilitates a shortcut connection that performs element-wise addition to blend the input and output. It enhances the flow of gradients and reduces the vanishing gradient problem which can be expressed as:
\begin{equation}
    M_k = S + M_{k-1},
\end{equation}
where \(M_k\) and \(M_{k-1}\) are the input and output of the $K$th MSRB and $S$ is the output of the blocks inside the MSRB.

\noindent\textbf{Hierarchical Feature Fusion Structure (HFFS)}\ \ \ \
This structure aims to preserve and enhance the features extracted across the MSRBs. As the network depth increases, it is imperative to maintain the fidelity of the input characteristics throughout the network. This process's output $F_{HFFS}$ can be formulated as:
\begin{equation}
    F_{HFFS} = \mathbf{W} * [M_0, M_1, \ldots, M_K] + b.
\end{equation}
Here, \(M_0\) represents the initial feature maps produced by the first convolutional layer, with subsequent \(M_i\)($i\neq 0$) denoting the outputs from each respective MSRB. $\mathbf{W}$ and $b$ are the weights and bias respectively.

\noindent\textbf{MSRN Image Reconstruction Stage}\ \ \ \
In the final stage, the image reconstruction module transforms the feature-enriched data back into high-resolution output using advanced upscaling techniques without initial upsampling. This approach integrates the PixelShuffle~\cite{shi2016real} method to efficiently upscale the image by directly rearranging the output tensor to a higher resolution. This technique significantly minimises the introduction of redundant information typically associated with traditional upscaling methods like bi-cubic interpolation.

In the rendering stage, we employ the MSRN to render super-resolution images. The network $F_\theta$ encapsulates the entire transformation process from $p^{(\text{LR})}$ to $p^{(\text{SR})}$ can be formulated as:
\begin{equation}
    p^{(SR)} = F_\theta(p^{(LR)}).
\end{equation}
\subsection{Training Losses}
Following 3DGS~\cite{kerbl20233d}, the loss function incorporates both L1 and D-SSIM terms. Also, our loss function contains our P-SSIM loss which is mentioned above and a total variant regulariser. The L1 term quantifies the absolute discrepancies between predictions and actual targets. The $L_1$ loss for the rendered colours is mathematically formulated as:
\begin{equation}
    L_1 = \frac{1}{B} \sum_{i=1}^B \sum_j |p_{i,j} - \hat{p}_{i,j}|,
\end{equation}
where \( p_{i,j} \) and \( \hat{p}_{i,j} \) are the pixels of the ground truth image $G_i$ and the rendered image $R_i$, and \( B \) represents the total number of pixels in the image. We also employ the Structural Similarity Index Measure loss and it is a loss term specifically designed to significantly enhance the perceptual quality of digital images and videos. By focusing on the structural discrepancies between target and rendered images, this metric provides a superior and more perceptually relevant evaluation compared to traditional pixel-based metrics. The D-SSIM loss is defined as:
\begin{equation}
L_{\text{D-SSIM}} = 1 - \text{SSIM}(p_{i,j}, \hat{p}_{i,j}),
\end{equation}
where SSIM~\cite{wang2004image} denotes the structural similarity between the target image and the rendered image.

To enhance image sharpness and mitigate excessive smoothness in the rendered image, our framework incorporates total variation (TV) regularisation. This regularisation technique is instrumental in preserving high-frequency details while maintaining overall image smoothness. The total variation loss is calculated by measuring the sum of absolute differences between adjacent pixels $\hat{p}$ in both horizontal and vertical directions in the rendered image. This approach emphasises maintaining edge sharpness by penalising large gradients in the image:
\begin{equation}
L_{tv} = \frac{\sum_{i, j} \left( |\hat{p}_{i,j} - \hat{p}_{i+1,j}| + |\hat{p}_{i,j} - \hat{p}_{i,j+1}| \right)}{\hat{c} \cdot h \cdot w}.  
\end{equation}
Here, \( \hat{p}_{i,j} \) represents the pixel intensity at position \( (i, j) \) in the rendered image $R_i$. The normalisation by \( \hat{c} \cdot h \cdot w \) (where \( \hat{c} \), \( h \), and \( w \) are the number of channels, height, and width of the image tensor, respectively) ensures that the loss calculation is scale-invariant with respect to the size of the image. The StructGS total loss function amalgamates the individual loss elements into a weighted sum which enhances the quality of reconstruction across diverse dimensions:
\begin{equation}
L_{\text{Total}} = \begin{cases}
    (1 - \lambda)L_{1} + \lambda L_{\text{D-SSIM}} +  \beta L_{tv},\quad \text{iterations} \leq k,\\
    (1 - \lambda)L_{1} + \lambda L_{\text{P-SSIM}} + \gamma L_{tv},\quad\text{otherwise,}
\end{cases}
\end{equation}
where $k$ denotes the iteration where we start to employ the P-SSIM loss. The \( \lambda \) and \( \beta \) are the respective weights assigned to the loss components. Through experiments, we set $k = 25000$, \( \lambda \) as 0.5, \( \beta \) as 0.04 and $\gamma$ as 0.02.
\section{Experiments}
\label{sec:experiment}
\subsection{Dataset and Metrics}
In our experimental setup, we assessed the performance of our model using 18 scenes sourced from various publicly available datasets. Specifically, we evaluated data tested in Scaffold-GS~\cite{lu2024scaffold}, Mip-Splatting~\cite{yu2024mip} and 3DGS~\cite{kerbl20233d} because they are our primary baselines. This included seven scenes from Mip-NeRF 360~\cite{barron2022mip}, two scenes each from Tanks \& Temples~\cite{knapitsch2017tanks} and DeepBlending~\cite{hedman2018deep}, six scenes from BungeeNeRF~\cite{xiangli2022bungeenerf} and one scene labelled ``Villa''. The ``Villa'' scene is a self-processed dataset. These scenes were carefully selected to encapsulate a range of contents captured at multiple levels of detail (LODs), showcasing our model's capabilities in view-adaptive rendering. They also span a diverse array of environments, covering both indoor and outdoor settings and incorporating large-scale scenes to demonstrate the scalability and versatility of our approach.

To quantitatively evaluate our model, we employed three widely-used metrics: Peak Signal-to-Noise Ratio (PSNR), Structural Similarity Index Measure (SSIM)~\cite{wang2004image}, and Learned Perceptual Image Patch Similarity (LPIPS)~\cite{zhang2018unreasonable}. Additionally, we explored the model's performance through progressive training across these complex 3D scenarios, allowing us to further validate and refine our approach based on detailed feedback and metric scores.

\begin{table*}
\centering
\caption{\textbf{Quantitative Comparison Results on the Mip-NeRF 360~\cite{barron2022mip}, Tanks\&Temples~\cite{knapitsch2017tanks} and DeepBlending~\cite{hedman2018deep} Datasets.} Our method outperforms baselines and SOTA methods~\cite{lu2024scaffold, yu2024mip}. The competing metrics are sourced from the respective papers (except for Mip-Splatting).}
\label{tab:mip_1}
\begin{tabular}{lccc|ccc|ccc}
\hline
& \multicolumn{3}{c|}{Mip-NeRF 360} & \multicolumn{3}{c|}{Tanks\&Temples} & \multicolumn{3}{c}{Deep Blending} \\
& PSNR $\uparrow$ & SSIM $\uparrow$ & LPIPS $\downarrow$ & PSNR $\uparrow$ & SSIM $\uparrow$ & LPIPS $\downarrow$ & PSNR $\uparrow$ & SSIM $\uparrow$ & LPIPS $\downarrow$  \\
\hline
Instant-NGP & 26.43 & 0.725 & 0.339 & 21.72 & 0.723 & 0.330 & 23.62 & 0.797 & 0.423 \\
Plenoxels & 23.62 & 0.670 & 0.443 & 21.08 & 0.719 & 0.379 & 23.06 & 0.795 & 0.510 \\
Mip-NeRF 360 & 29.23 & 0.844 & 0.207 & 22.22 & 0.759 & 0.257 & 29.40 & 0.901 & 0.245 \\
\hline
3DGS & 28.69 & 0.870 & 0.182 & 23.14 & 0.841 & 0.183 & 29.41 & 0.903 & 0.243 \\
Mip-Splatting & 29.09 & 0.871 & 0.184 & 23.87 & 0.851 & 0.176 & 29.70 & 0.905 & 0.243 \\
Scaffold-GS & 28.84 & 0.848 & 0.220 & 23.96 & 0.853 & 0.177 & 30.21 & 0.906 & 0.254\\
\hline
Ours (without MSRN) & 29.21 & 0.881 & 0.155 & 23.83 & 0.863 & 0.141 & 30.19 & 0.905 & 0.244 \\
Ours (Full) & \textbf{30.69} & \textbf{0.928} & \textbf{0.036} & \textbf{24.55} & \textbf{0.893} & \textbf{0.051} & \textbf{30.28} & \textbf{0.908} & \textbf{0.079} \\
\hline
\end{tabular}
\end{table*}

\begin{table*}
\centering
\caption{\textbf{Quantitative Comparison Results on the BungeeNeRF~\cite{xiangli2022bungeenerf} and Villa Datasets.} Our method outperforms baselines and SOTA methods~\cite{lu2024scaffold, yu2024mip}.}
\label{tab:mip_2}
\begin{tabular}{lccc|ccc}
\hline
& \multicolumn{3}{c|}{BungeeNeRF} & \multicolumn{3}{c}{Villa} \\
& PSNR $\uparrow$ & SSIM $\uparrow$ & LPIPS $\downarrow$ & PSNR $\uparrow$ & SSIM $\uparrow$ & LPIPS $\downarrow$ \\
\hline
3DGS & 24.89 & 0.827 & 0.211 & 25.40 & 0.862 & 0.160  \\
Mip-Splatting & 28.25 & 0.919 & 0.097 & 25.74 & 0.869 & 0.174 \\
Scaffold-GS & 27.03 & 0.899 & 0.101 & 25.81 & 0.869 & 0.164 \\
\hline
Ours (without MSRN) & 27.99 & 0.917 & 0.091 & 25.41 & 0.870 & 0.129 \\
Ours (Full) & \textbf{31.01} & \textbf{0.960} & \textbf{0.025} & \textbf{25.99} & \textbf{0.912} & \textbf{0.043} \\
\hline
\end{tabular}
\end{table*}

\begin{figure*}[!t]
    \centering
    \includegraphics[width=\textwidth]{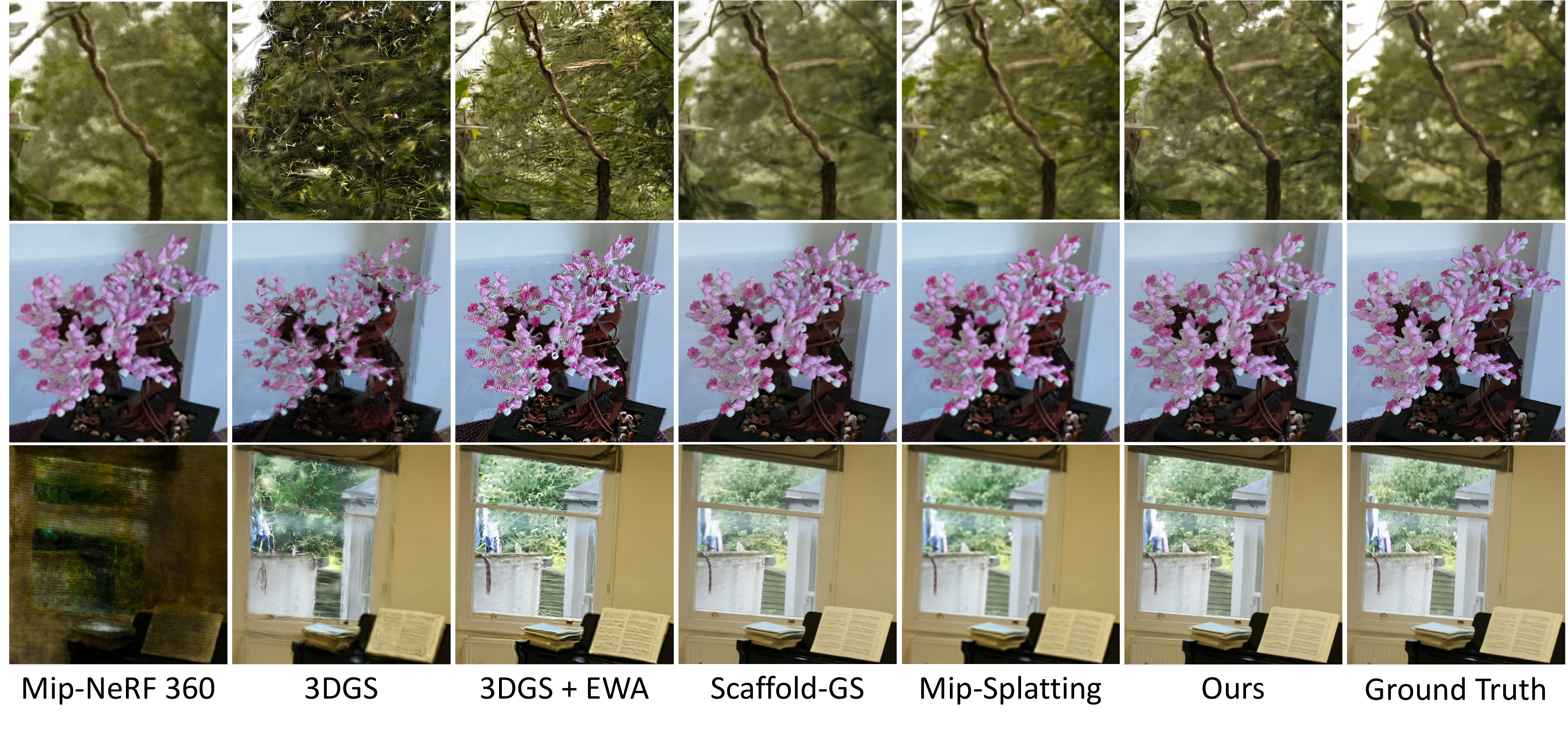}
    \vspace{-2em}
    \caption{\textbf{Qualitative Comparison Results on the Mip-NeRF 360 Dataset~\cite{barron2022mip}.} These models were trained on images with a resolution of 1.6k and we simulated the zoom-in situation. Unlike previous approaches, our model attains a higher degree of accuracy and detail than other models, rendering images that closely match the ground truth.}
    \label{fig:mip-nerf-360}
\end{figure*}

\begin{figure*}
    \centering
    \includegraphics[width=\textwidth]{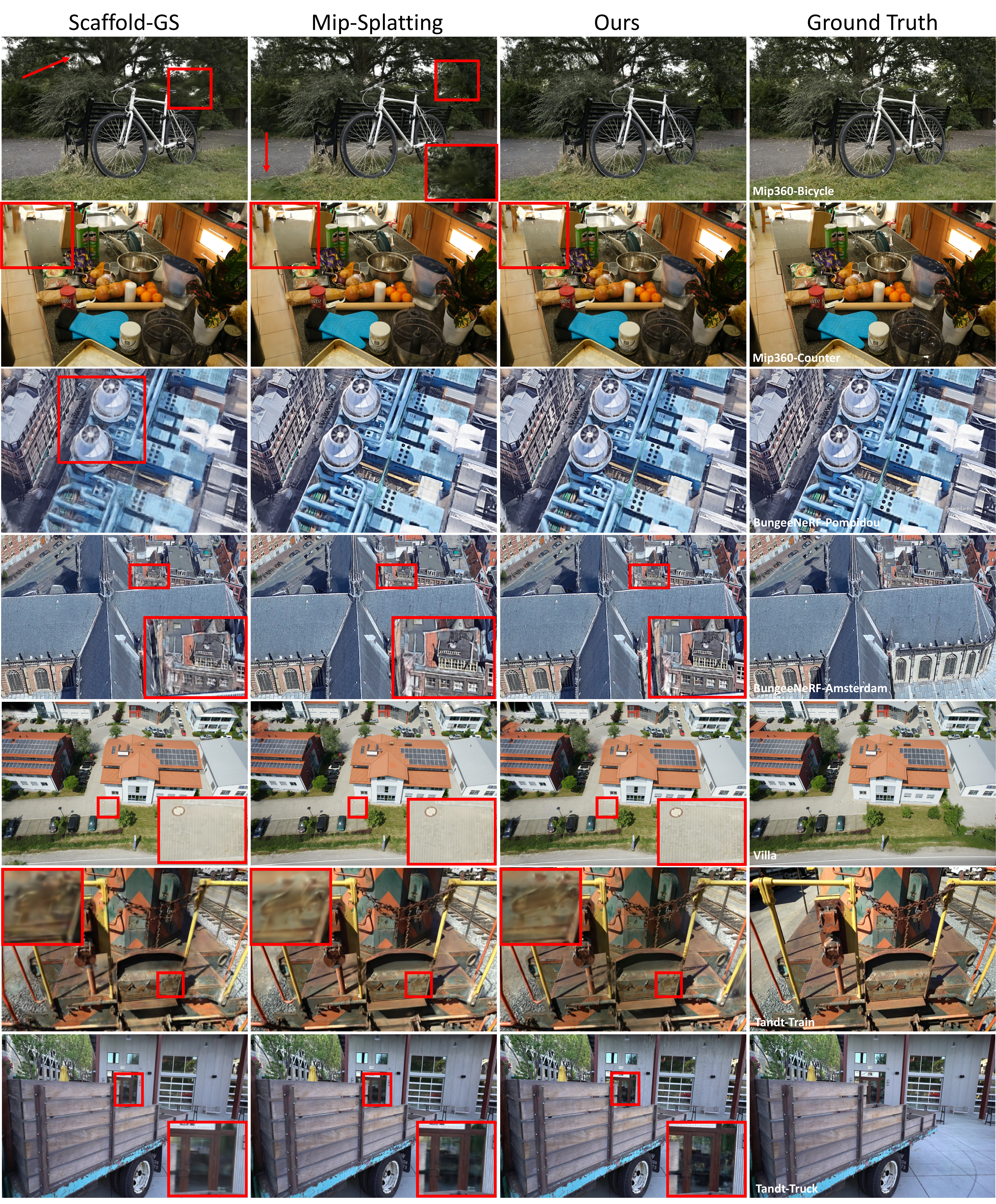}
    \caption{\textbf{Qualitative Comparison Results across diverse datasets~\cite{barron2022mip, xiangli2022bungeenerf, knapitsch2017tanks}.} The red boxes and arrows highlight artifacts rendered by state-of-the-art models~\cite{yu2024mip, lu2024scaffold}, which cause certain areas of the rendered images to appear blurry. The results above demonstrate that our model outperforms state-of-the-art models, achieving superior performance in rendering details with significantly fewer artifacts.}
    \label{fig:compare_2}
\end{figure*}
\subsection{Baselines and Implementation}
For our experimental benchmarking, we selected Scaffold-GS~\cite{lu2024scaffold}, Mip-Splatting~\cite{yu2024mip} and 3DGS~\cite{kerbl20233d} as our primary baselines due to their state-of-the-art (SOTA) performance in 3D reconstruction and novel view synthesis. To ensure a fair comparison, we trained our model along with all selected baselines for 30k iterations. Additionally, we included results from Mip-NeRF 360~\cite{barron2022mip}, Instant-NGP~\cite{muller2022instant}, Plenoxels~\cite{fridovich2022plenoxels} and 3DGS enhanced with EWA~\cite{zwicker2001ewa} to provide a comprehensive overview of current capabilities in the field.

In our implementation, we set \(P = 10\), where \(P\) represents the number of stochastic patch pairs assessed during training, and \(k = 25,000\) to control the transition to Patch-based Structural Similarity Index Measure (P-SSIM). We structured the training process around three primary loss weights: \(\lambda = 0.5\), \(\beta = 0.04\), and \(\gamma = 0.02\). The reason why we select $\lambda$ as 0.5 is because we aim to increase the importance of P-SSIM. The selection of weights for TV loss is motivated by a desire to enhance the effect of TV Loss for the initial $k$ iterations, to suppress noise and irregularities in the model more effectively. Following this phase, we aim to emphasise the importance of P-SSIM by reducing the weight of TV loss. This reduction is also intended to preserve some fine structures and details. Furthermore, our refinement process includes a pruning step where an anchor is eliminated if the accumulated opacity of its Gaussians falls below 0.005, ensuring that only significant features contribute to the final model structure.
\subsection{Results Comparison}
In evaluating the efficacy of our method, we conducted comparisons against several baselines including Scaffold-GS, Mip-Splatting, 3D-GS, Mip-NeRF360, Instant-NGP, and Plenoxels using real-world datasets. Qualitative results are presented in Tab.~\ref{tab:mip_1} and ~\ref{tab:mip_2}. The performance metrics for Mip-NeRF 360~\cite{barron2022mip}, Instant-NGP~\cite{muller2022instant} and Plenoxels~\cite{fridovich2022plenoxels} are consistent with those presented in the previous study~\cite{kerbl20233d}. We conducted the training for Scaffold-GS~\cite{lu2024scaffold} and Mip-Splatting~\cite{yu2024mip} using default settings provided by their respective implementations. Notably, our model outperformed all other models, including state-of-the-art (SOTA) methods~\cite{lu2024scaffold, yu2024mip}, across all dataset scenes. Despite the removal of the Multi-scale Residual Network (MSRN) from our model, it still outperformed SOTA 3D-GS-based methods~\cite{kerbl20233d, lu2024scaffold, yu2024mip} in the Mip-NeRF 360 dataset~\cite{barron2022mip} and achieved comparable results in other datasets as detailed in Tab.~\ref{tab:mip_1} and ~\ref{tab:mip_2}.

Fig.~\ref{fig:mip-nerf-360} illustrates the zoom-in performance comparison of our model with the baselines, where our model visibly surpasses the visual quality of SOTA~\cite{lu2024scaffold, yu2024mip}. Both 3DGS~\cite{kerbl20233d} and 3DGS + EWA~\cite{zwicker2001ewa} exhibited notable erosion artifacts due to dilation operations and some high-frequency artifacts. In comparison to Scaffold-GS~\cite{lu2024scaffold} and Mip-Splatting~\cite{yu2024mip}, our model rendered images with clearer details and higher resolution, while Scaffold-GS and Mip-Splatting exhibited various degrees of blurry artifacts. Our model avoids such artifacts and closely matches the ground truth. Fig.~\ref{fig:compare_2} provides a more detailed view of our model’s rendering quality compared to Scaffold-GS~\cite{lu2024scaffold} and Mip-Splatting~\cite{yu2024mip}. During training, it was observed that Scaffold-GS and Mip-Splatting often produced blurry outcomes and Scaffold-GS was particularly susceptible to high-frequency artifacts. As depicted in Fig.~\ref{fig:compare_2}, which represents different scenes from various datasets, our model demonstrates its ability to preserve superior detail. Even when zooming into specific areas, it maintains complete and crisp details. In contrast, Mip-Splatting and Scaffold-GS frequently lose detail and exhibit blurriness.
\subsection{Ablation Study}
\begin{table*}
\centering
\caption{\textbf{Ablation Study of Components in Our Model.} We present quantitative results for the Bicycle, Counter and Kitchen scenes from the Mip-NeRF 360 dataset~\cite{barron2022mip} and Tanks\&Temples dataset~\cite{knapitsch2017tanks}. All scenes are trained for 30k iterations and the input image is at a resolution of 1.6k.}
\label{tab:abltation}
\begin{tabular}{lccc|ccc|ccc|ccc}
\hline
& \multicolumn{3}{c|}{Bicycle} & \multicolumn{3}{c|}{Counter} & \multicolumn{3}{c|}{Kitchen} & \multicolumn{3}{c}{Tanks\&Temples} \\
& PSNR $\uparrow$ & SSIM $\uparrow$ & LPIPS $\downarrow$ & PSNR $\uparrow$ & SSIM $\uparrow$ & LPIPS $\downarrow$ & PSNR $\uparrow$ & SSIM $\uparrow$ & LPIPS $\downarrow$ & PSNR $\uparrow$ & SSIM $\uparrow$ & LPIPS $\downarrow$ \\
\hline
None & 25.13 & 0.747 & 0.245 & 29.13 & 0.915 & 0.183 & 31.47 & 0.932 & 0.116 & 23.33 & 0.845 & 0.179 \\
w/ $L_{P-SSIM}$ & 25.64 & 0.785 & 0.178 & 29.40 & 0.920 & 0.161 & 31.90 & 0.937 & 0.106 & 23.79 & 0.855 & 0.151 \\
w/ $L_{tv}$ & 25.18 & 0.750 & 0.229 & 29.21 & 0.916 & 0.177 & 31.55 & 0.933 & 0.110 & 23.48 & 0.850 & 0.163 \\
w/ MSRN & 27.51 & 0.832 & 0.091 & 29.66 & 0.927 & 0.087 & 32.11 & 0.954 & 0.044 & 24.11 & 0.878 & 0.099 \\
Full & \textbf{28.07} & \textbf{0.884} & \textbf{0.047} & \textbf{30.13} & \textbf{0.939} & \textbf{0.045} & \textbf{33.37} & \textbf{0.965} & \textbf{0.022} & \textbf{24.55} & \textbf{0.893} & \textbf{0.051} \\
\hline
\end{tabular}
\end{table*}

\subsubsection{Loss terms and Regulariser}
We conducted an ablation study to evaluate the impact of the designed loss terms which are \( L_{P-SSIM} \) and \( L_{tv} \). We trained our model on three scenes from the Mip-NeRF 360 dataset~\cite{barron2022mip} and two scenes from the Tanks\&Temples dataset~\cite{knapitsch2017tanks}, including the ``Bicycle'', ``Counter'', ``Kitchen'', ``Train'' and ``Truck'' scenes for 30k iterations. They are the most commonly used scenes in 3D reconstruction. As indicated in Tab.~\ref{tab:abltation}, the inclusion of \( L_{P-SSIM} \) and \( L_{tv} \) in our original model led to improvements across all performance metrics. Notably, \( L_{P-SSIM} \) had the most significant impact due to its ability to help the model capture more non-local structural information with stochastic characteristics during training. \( L_{tv} \), serving as an auxiliary regulariser, further enhanced the model's performance by promoting smoothness and reducing artifacts.
\begin{figure*}[!t]
    \centering
    \includegraphics[width=1\textwidth]{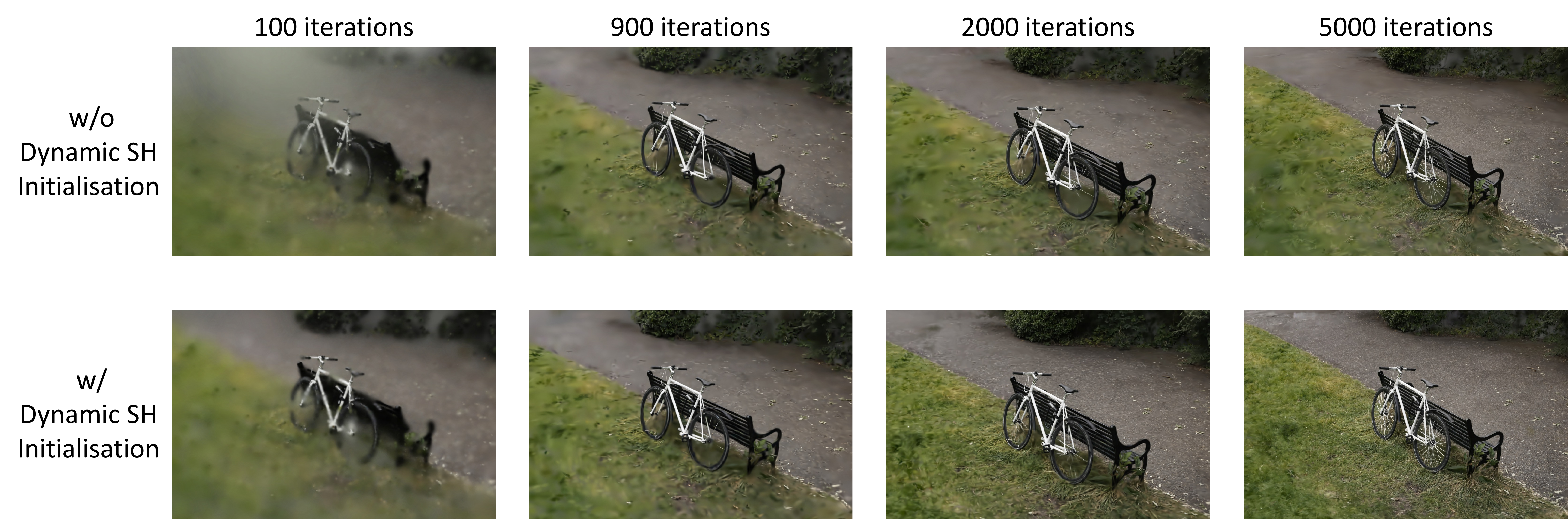}
    \caption{\textbf{Ablation of Dynamic Spherical Harmonics Initialisation.} We present an ablation study of the training progression of the bicycle scene~\cite{barron2022mip}. The result shows that this strategy improves the training quality in the early iterations.}
    \label{fig:abltaion_sh}
\end{figure*}

\subsubsection{Dynamic Spherical Harmonics Initialisation}
We evaluated our method for the dynamic adjustment of spherical harmonics initialisation, as described in Section~\ref{sec:experiment}. In Fig.~\ref{fig:abltaion_sh}, we present the training progression of the Bicycle scene with and without applying this strategy, showing results at 100, 900, 2000 and 5000 training iterations. It is evident that the model utilising our dynamic initialisation strategy outperforms the standard approach in early training phases. This advantage is particularly noticeable in aspects such as richer colours and more realistic details. Our method addresses the issue of high-degree spherical harmonics not being initialised and engaged during early training iterations. By dynamically adjusting the initialisation of spherical harmonics, our model achieves better performance in the initial stages of training.
\subsubsection{Multi-scale Residual Network}
We assessed our model both with and without the implementation of the Multi-scale Residual Network. The results of this comparison are presented in Tab.~\ref{tab:mip_1}, Tab.~\ref{tab:mip_2} and Tab.~\ref{tab:abltation}, where we display the metric scores for all datasets, including scenarios where the Multi-scale Residual Network (MSRN) was removed. As evident from these tables, the inclusion of the Multi-scale Residual Network (MSRN) significantly enhances performance across various datasets. This enhancement is attributed to the network's capability to render higher resolution images during the rendering phase of our model, which in turn improves the expressiveness of the pixels.
\subsection{Limitation and Future Work}
Our model leverages a patch-based SSIM loss, which has significantly improved performance; however, this approach requires the image to be segmented into numerous patches to compute SSIM for each segment, thereby increasing computational resource demands and extending overall training duration. Currently, we apply the P-SSIM loss only to RGB values. In the future, we may extend the application of structural similarity metrics like S3IM to non-RGB losses, such as those involving depth information. Another limitation involves our method of Dynamic Adjustment of Spherical Harmonics Initialisation, which heavily relies on initial estimates of opacity and RGB colours. Poor initial estimates can adversely affect the efficacy of this strategy. We are considering future enhancements that would allow dynamic adjustments of spherical harmonic values throughout the training process, making it more adaptive rather than confined to the initialisation phase. Lastly, the use of the Multi-scale Residual Network (MSRN) in the rendering phase increases rendering times by approximately 30\%. Despite this, it still supports relatively fast rendering. Future iterations of our model may integrate MSRN with 3DGS during training, applying super-resolution information directly to enhance both training efficiency and rendering quality.
\section{Conclusion}
\label{sec:conclusion}
In this paper, we introduced StructGS, an enhanced model based on the original 3D Gaussian Splatting framework. Our approach includes three main innovations: Non-Local Structural Information P-SSIM, Dynamic Adjustment of Spherical Harmonics Initialisation and the incorporation of a Multi-scale Residual Network (MSRN) during the rendering phase. These innovations aim to address the limitations of 3DGS-based models in capturing stochastic, non-local structural information during training and to improve the early training performance as well as the rendering quality of the model. Experimental results demonstrate that StructGS significantly outperforms state-of-the-art (SOTA) models across various datasets, including indoor, outdoor and large-scale scenes. It delivers higher quality, higher resolution images with more detailed content and fewer high-frequency artifacts.
{\appendix[More Experiments Result]
In the appendix, we present the evaluation metrics for our baselines and our model across all scenes in the Mip-NeRF 360 dataset~\cite{barron2022mip}. It is evident that our model outperforms the state-of-the-art (SOTA) models based on 3DGS~\cite{yu2024mip,lu2024scaffold} in terms of PSNR and SSIM~\cite{wang2004image} metrics in almost all scenes. This is true even without employing MSRN for super-resolution rendering. In a few scenes, the results are comparable. Table~\ref{tab:mip-LPIPS} illustrates that our model's images, whether MSRN is used or not, significantly surpass all other models in LPIPS~\cite{zhang2018unreasonable} scores.

\begin{table}[ht]
\centering
\caption{SSIM of baselines and our method for Mip-NeRF 360 dataset~\cite{barron2022mip}.}
\label{tab:mip-SSIM}
\begin{adjustbox}{width=3.45in,center}
\begin{tabular}{l|ccccccc}
\hline
\multicolumn{1}{c|}{Method \textbar\ Scenes} & bicycle & garden & stump & room & counter & kitchen & bonsai \\
\hline
Instant-NGP & 0.491 & 0.649 & 0.574 & 0.855 & 0.798 & 0.818 & 0.890 \\
Plenoxels & 0.496 & 0.606 & 0.523 & 0.842 & 0.759 & 0.648 & 0.814 \\
Mip-NeRF 360 & 0.685 & 0.813 & 0.744 & 0.913 & 0.894 & 0.920 & 0.941 \\
3DGS & 0.771 & 0.868 & 0.775 & 0.914 & 0.905 & 0.922 & 0.938 \\
Scaffold-GS & 0.705 & 0.842 & 0.784 & 0.925 & 0.914 & 0.928 & 0.946 \\
Mip-Splatting & 0.747 & 0.858 & 0.770 & 0.927 & 0.915 & 0.932 & 0.947 \\ 
\hline
Ours (without MSRN) & 0.785 & 0.865 & 0.783 & 0.927 & 0.920 & 0.937 & 0.951 \\
Ours (Full) & \textbf{0.884} & \textbf{0.935} & \textbf{0.854} & \textbf{0.951} & \textbf{0.939} & \textbf{0.965} & \textbf{0.968} \\
\hline
\end{tabular}
\end{adjustbox}
\end{table}

\begin{table}[ht]
\centering
\caption{PSNR of baselines and our method for Mip-NeRF 360 dataset~\cite{barron2022mip}.}
\label{tab:mip-PSNR}
\begin{adjustbox}{width=3.45in,center}
\begin{tabular}{l|ccccccc}
\hline
\multicolumn{1}{c|}{Method \textbar\ Scenes} & bicycle & garden & stump & room & counter & kitchen & bonsai \\
\hline
Instant-NGP & 22.19 & 24.60 & 23.63 & 29.27 & 26.44 & 28.55 & 30.34 \\
Plenoxels  & 21.91 & 23.49 & 20.66 & 27.59 & 23.62 & 23.42 & 24.67 \\
Mip-NeRF 360 & 24.37 & 26.98 & 26.40 & 31.63 & 29.55 & 32.23 & \textbf{33.46} \\
3DGS & 25.25 & 27.41 & 26.55 & 30.63 & 28.70 & 30.32 & 31.98 \\
Scaffold-GS  & 24.50 & 27.17 & 26.27 & 31.93 & 29.34 & 31.30 & 32.70 \\
Mip-Splatting & 25.13 & 27.38 & 26.64 & 31.50 & 29.13 & 31.47 & 32.39 \\ 
\hline
Ours (without MSRN) & 25.64 & 27.18 & 26.79 & 31.05 & 29.40 & 31.90 & 32.48 \\
Ours (Full) & \textbf{28.07} & \textbf{29.60} & \textbf{28.18} & \textbf{32.10} & \textbf{30.13} & \textbf{33.37} & 33.41 \\
\hline
\end{tabular}
\end{adjustbox}
\end{table}

\begin{table}[ht]
\centering
\caption{LPIPS of baselines and our method for Mip-NeRF 360 dataset~\cite{barron2022mip}.}
\label{tab:mip-LPIPS}
\begin{adjustbox}{width=3.45in,center}
\begin{tabular}{l|ccccccc}
\hline
\multicolumn{1}{c|}{Method \textbar\ Scenes} & bicycle & garden & stump & room & counter & kitchen & bonsai \\
\hline
Instant-NGP  & 0.487 & 0.312 & 0.450 & 0.301 & 0.342 & 0.254 & 0.227 \\
Plenoxels  & 0.506 & 0.386 & 0.503 & 0.419 & 0.441 & 0.447 & 0.398 \\
Mip-NeRF 360 & 0.301 & 0.170 & 0.261 & 0.211 & 0.204 & 0.127 & 0.176 \\
3DGS  & 0.205 & 0.103 & 0.210 & 0.220 & 0.204 & 0.129 & 0.205 \\
Scaffold-GS  & 0.306 & 0.146 & 0.284 & 0.202 & 0.191 & 0.126 & 0.185 \\
Mip-Splatting & 0.245 & 0.123 & 0.243 & 0.197 & 0.183 & 0.116 & 0.180 \\ 
\hline
Ours (without MSRN) & 0.178 & 0.104 & 0.201 & 0.178 & 0.161 & 0.106 & 0.156 \\
Ours (Full) & \textbf{0.047} & \textbf{0.026} & \textbf{0.057} & \textbf{0.040} & \textbf{0.045} & \textbf{0.022} & \textbf{0.018} \\
\hline
\end{tabular}
\end{adjustbox}
\end{table}


\bibliographystyle{IEEEtran}
\bibliography{IEEEabrv,references}

\begin{thebibliography}{10}
\providecommand{\url}[1]{#1}
\csname url@samestyle\endcsname
\providecommand{\newblock}{\relax}
\providecommand{\bibinfo}[2]{#2}
\providecommand{\BIBentrySTDinterwordspacing}{\spaceskip=0pt\relax}
\providecommand{\BIBentryALTinterwordstretchfactor}{4}
\providecommand{\BIBentryALTinterwordspacing}{\spaceskip=\fontdimen2\font plus
\BIBentryALTinterwordstretchfactor\fontdimen3\font minus \fontdimen4\font\relax}
\providecommand{\BIBforeignlanguage}[2]{{%
\expandafter\ifx\csname l@#1\endcsname\relax
\typeout{** WARNING: IEEEtran.bst: No hyphenation pattern has been}%
\typeout{** loaded for the language `#1'. Using the pattern for}%
\typeout{** the default language instead.}%
\else
\language=\csname l@#1\endcsname
\fi
#2}}
\providecommand{\BIBdecl}{\relax}
\BIBdecl

\bibitem{mildenhall2021nerf}
B.~Mildenhall, P.~P. Srinivasan, M.~Tancik, J.~T. Barron, R.~Ramamoorthi, and R.~Ng, ``Nerf: Representing scenes as neural radiance fields for view synthesis,'' \emph{Communications of the ACM}, vol.~65, no.~1, pp. 99--106, 2021.

\bibitem{barron2021mip}
J.~T. Barron, B.~Mildenhall, M.~Tancik, P.~Hedman, R.~Martin-Brualla, and P.~P. Srinivasan, ``Mip-nerf: A multiscale representation for anti-aliasing neural radiance fields,'' in \emph{Proceedings of the IEEE/CVF international conference on computer vision}, 2021, pp. 5855--5864.

\bibitem{barron2022mip}
J.~T. Barron, B.~Mildenhall, D.~Verbin, P.~P. Srinivasan, and P.~Hedman, ``Mip-nerf 360: Unbounded anti-aliased neural radiance fields,'' in \emph{Proceedings of the IEEE/CVF conference on computer vision and pattern recognition}, 2022, pp. 5470--5479.

\bibitem{muller2022instant}
T.~M{\"u}ller, A.~Evans, C.~Schied, and A.~Keller, ``Instant neural graphics primitives with a multiresolution hash encoding,'' \emph{ACM transactions on graphics (TOG)}, vol.~41, no.~4, pp. 1--15, 2022.

\bibitem{sun2022direct}
C.~Sun, M.~Sun, and H.-T. Chen, ``Direct voxel grid optimization: Super-fast convergence for radiance fields reconstruction,'' in \emph{Proceedings of the IEEE/CVF conference on computer vision and pattern recognition}, 2022, pp. 5459--5469.

\bibitem{kerbl20233d}
B.~Kerbl, G.~Kopanas, T.~Leimk{\"u}hler, and G.~Drettakis, ``3d gaussian splatting for real-time radiance field rendering.'' \emph{ACM Trans. Graph.}, vol.~42, no.~4, pp. 139--1, 2023.

\bibitem{schonberger2016structure}
J.~L. Schonberger and J.-M. Frahm, ``Structure-from-motion revisited,'' in \emph{Proceedings of the IEEE conference on computer vision and pattern recognition}, 2016, pp. 4104--4113.

\bibitem{yu2024mip}
Z.~Yu, A.~Chen, B.~Huang, T.~Sattler, and A.~Geiger, ``Mip-splatting: Alias-free 3d gaussian splatting,'' in \emph{Proceedings of the IEEE/CVF Conference on Computer Vision and Pattern Recognition}, 2024, pp. 19\,447--19\,456.

\bibitem{lu2024scaffold}
T.~Lu, M.~Yu, L.~Xu, Y.~Xiangli, L.~Wang, D.~Lin, and B.~Dai, ``Scaffold-gs: Structured 3d gaussians for view-adaptive rendering,'' in \emph{Proceedings of the IEEE/CVF Conference on Computer Vision and Pattern Recognition}, 2024, pp. 20\,654--20\,664.

\bibitem{xie2023s3im}
Z.~Xie, X.~Yang, Y.~Yang, Q.~Sun, Y.~Jiang, H.~Wang, Y.~Cai, and M.~Sun, ``S3im: Stochastic structural similarity and its unreasonable effectiveness for neural fields,'' in \emph{Proceedings of the IEEE/CVF International Conference on Computer Vision}, 2023, pp. 18\,024--18\,034.

\bibitem{li2018multi}
J.~Li, F.~Fang, K.~Mei, and G.~Zhang, ``Multi-scale residual network for image super-resolution,'' in \emph{Proceedings of the European conference on computer vision (ECCV)}, 2018, pp. 517--532.

\bibitem{debevec1998efficient}
P.~Debevec, Y.~Yu, and G.~Boshokov, ``Efficient view-dependent ibr with projective texture-mapping,'' in \emph{EG Rendering Workshop}, vol.~4, no.~11, 1998.

\bibitem{goesele2007multi}
M.~Goesele, N.~Snavely, B.~Curless, H.~Hoppe, and S.~M. Seitz, ``Multi-view stereo for community photo collections,'' in \emph{2007 IEEE 11th International Conference on Computer Vision}.\hskip 1em plus 0.5em minus 0.4em\relax IEEE, 2007, pp. 1--8.

\bibitem{schonberger2016pixelwise}
J.~L. Sch{\"o}nberger, E.~Zheng, J.-M. Frahm, and M.~Pollefeys, ``Pixelwise view selection for unstructured multi-view stereo,'' in \emph{Computer Vision--ECCV 2016: 14th European Conference, Amsterdam, The Netherlands, October 11-14, 2016, Proceedings, Part III 14}.\hskip 1em plus 0.5em minus 0.4em\relax Springer, 2016, pp. 501--518.

\bibitem{shum2000review}
H.~Shum and S.~B. Kang, ``Review of image-based rendering techniques,'' \emph{Visual Communications and Image Processing 2000}, vol. 4067, pp. 2--13, 2000.

\bibitem{tewari2022advances}
A.~Tewari, J.~Thies, B.~Mildenhall, P.~Srinivasan, E.~Tretschk, W.~Yifan, C.~Lassner, V.~Sitzmann, R.~Martin-Brualla, S.~Lombardi \emph{et~al.}, ``Advances in neural rendering,'' in \emph{Computer Graphics Forum}, vol.~41, no.~2.\hskip 1em plus 0.5em minus 0.4em\relax Wiley Online Library, 2022, pp. 703--735.

\bibitem{qin2022bullet}
H.~Qin, J.~Li, Y.~Jiang, Y.~Dai, S.~Hong, F.~Zhou, Z.~Wang, and T.~Yang, ``Bullet-time video synthesis based on virtual dynamic target axis,'' \emph{IEEE Transactions on Multimedia}, vol.~25, pp. 5178--5191, 2022.

\bibitem{alexiadis2012real}
D.~S. Alexiadis, D.~Zarpalas, and P.~Daras, ``Real-time, full 3-d reconstruction of moving foreground objects from multiple consumer depth cameras,'' \emph{IEEE Transactions on Multimedia}, vol.~15, no.~2, pp. 339--358, 2012.

\bibitem{park2019deepsdf}
J.~J. Park, P.~Florence, J.~Straub, R.~Newcombe, and S.~Lovegrove, ``Deepsdf: Learning continuous signed distance functions for shape representation,'' in \emph{Proceedings of the IEEE/CVF conference on computer vision and pattern recognition}, 2019, pp. 165--174.

\bibitem{wang2021neus}
P.~Wang, L.~Liu, Y.~Liu, C.~Theobalt, T.~Komura, and W.~Wang, ``Neus: Learning neural implicit surfaces by volume rendering for multi-view reconstruction,'' \emph{NeurIPS}, 2021.

\bibitem{wang2023neus2}
Y.~Wang, Q.~Han, M.~Habermann, K.~Daniilidis, C.~Theobalt, and L.~Liu, ``Neus2: Fast learning of neural implicit surfaces for multi-view reconstruction,'' in \emph{Proceedings of the IEEE/CVF International Conference on Computer Vision}, 2023, pp. 3295--3306.

\bibitem{huang2023natural}
Y.~Huang, F.~Juefei-Xu, Q.~Guo, G.~Pu, and Y.~Liu, ``Natural \& adversarial bokeh rendering via circle-of-confusion predictive network,'' \emph{IEEE Transactions on Multimedia}, 2023.

\bibitem{chen2022tensorf}
A.~Chen, Z.~Xu, A.~Geiger, J.~Yu, and H.~Su, ``Tensorf: Tensorial radiance fields,'' in \emph{European conference on computer vision}.\hskip 1em plus 0.5em minus 0.4em\relax Springer, 2022, pp. 333--350.

\bibitem{huang2024efficient}
Z.~Huang, S.~M. Erfani, S.~Lu, and M.~Gong, ``Efficient neural implicit representation for 3d human reconstruction,'' \emph{Pattern Recognition}, vol. 156, p. 110758, 2024.

\bibitem{fridovich2022plenoxels}
S.~Fridovich-Keil, A.~Yu, M.~Tancik, Q.~Chen, B.~Recht, and A.~Kanazawa, ``Plenoxels: Radiance fields without neural networks,'' in \emph{Proceedings of the IEEE/CVF conference on computer vision and pattern recognition}, 2022, pp. 5501--5510.

\bibitem{yu2021plenoctrees}
A.~Yu, R.~Li, M.~Tancik, H.~Li, R.~Ng, and A.~Kanazawa, ``Plenoctrees for real-time rendering of neural radiance fields,'' in \emph{Proceedings of the IEEE/CVF International Conference on Computer Vision}, 2021, pp. 5752--5761.

\bibitem{liao2022kitti}
Y.~Liao, J.~Xie, and A.~Geiger, ``Kitti-360: A novel dataset and benchmarks for urban scene understanding in 2d and 3d,'' \emph{IEEE Transactions on Pattern Analysis and Machine Intelligence}, vol.~45, no.~3, pp. 3292--3310, 2022.

\bibitem{schutz2019real}
M.~Sch{\"u}tz, K.~Kr{\"o}sl, and M.~Wimmer, ``Real-time continuous level of detail rendering of point clouds,'' in \emph{2019 IEEE Conference on Virtual Reality and 3D User Interfaces (VR)}.\hskip 1em plus 0.5em minus 0.4em\relax IEEE, 2019, pp. 103--110.

\bibitem{ruckert2022adop}
D.~R{\"u}ckert, L.~Franke, and M.~Stamminger, ``Adop: Approximate differentiable one-pixel point rendering,'' \emph{ACM Transactions on Graphics (ToG)}, vol.~41, no.~4, pp. 1--14, 2022.

\bibitem{kopanas2021point}
G.~Kopanas, J.~Philip, T.~Leimk{\"u}hler, and G.~Drettakis, ``Point-based neural rendering with per-view optimization,'' in \emph{Computer Graphics Forum}, vol.~40, no.~4.\hskip 1em plus 0.5em minus 0.4em\relax Wiley Online Library, 2021, pp. 29--43.

\bibitem{wiles2020synsin}
O.~Wiles, G.~Gkioxari, R.~Szeliski, and J.~Johnson, ``Synsin: End-to-end view synthesis from a single image,'' in \emph{Proceedings of the IEEE/CVF conference on computer vision and pattern recognition}, 2020, pp. 7467--7477.

\bibitem{arvanitis2021broad}
G.~Arvanitis, E.~I. Zacharaki, L.~V{\'a}{\^s}a, and K.~Moustakas, ``Broad-to-narrow registration and identification of 3d objects in partially scanned and cluttered point clouds,'' \emph{IEEE Transactions on Multimedia}, vol.~24, pp. 2230--2245, 2021.

\bibitem{jiang2024gs}
Y.~Jiang, J.~Li, H.~Qin, Y.~Dai, J.~Liu, G.~Zhang, C.~Zhang, and T.~Yang, ``Gs-sfs: Joint gaussian splatting and shape-from-silhouette for multiple human reconstruction in large-scale sports scenes,'' \emph{IEEE Transactions on Multimedia}, 2024.

\bibitem{shi2016real}
W.~Shi, J.~Caballero, F.~Husz{\'a}r, J.~Totz, A.~P. Aitken, R.~Bishop, D.~Rueckert, and Z.~Wang, ``Real-time single image and video super-resolution using an efficient sub-pixel convolutional neural network,'' in \emph{Proceedings of the IEEE conference on computer vision and pattern recognition}, 2016, pp. 1874--1883.

\bibitem{wang2004image}
Z.~Wang, A.~C. Bovik, H.~R. Sheikh, and E.~P. Simoncelli, ``Image quality assessment: from error visibility to structural similarity,'' \emph{IEEE transactions on image processing}, vol.~13, no.~4, pp. 600--612, 2004.

\bibitem{knapitsch2017tanks}
A.~Knapitsch, J.~Park, Q.-Y. Zhou, and V.~Koltun, ``Tanks and temples: Benchmarking large-scale scene reconstruction,'' \emph{ACM Transactions on Graphics (ToG)}, vol.~36, no.~4, pp. 1--13, 2017.

\bibitem{hedman2018deep}
P.~Hedman, J.~Philip, T.~Price, J.-M. Frahm, G.~Drettakis, and G.~Brostow, ``Deep blending for free-viewpoint image-based rendering,'' \emph{ACM Transactions on Graphics (ToG)}, vol.~37, no.~6, pp. 1--15, 2018.

\bibitem{xiangli2022bungeenerf}
Y.~Xiangli, L.~Xu, X.~Pan, N.~Zhao, A.~Rao, C.~Theobalt, B.~Dai, and D.~Lin, ``Bungeenerf: Progressive neural radiance field for extreme multi-scale scene rendering,'' in \emph{European conference on computer vision}.\hskip 1em plus 0.5em minus 0.4em\relax Springer, 2022, pp. 106--122.

\bibitem{zhang2018unreasonable}
R.~Zhang, P.~Isola, A.~A. Efros, E.~Shechtman, and O.~Wang, ``The unreasonable effectiveness of deep features as a perceptual metric,'' in \emph{Proceedings of the IEEE conference on computer vision and pattern recognition}, 2018, pp. 586--595.

\bibitem{zwicker2001ewa}
M.~Zwicker, H.~Pfister, J.~Van~Baar, and M.~Gross, ``Ewa volume splatting,'' in \emph{Proceedings Visualization, 2001. VIS'01.}\hskip 1em plus 0.5em minus 0.4em\relax IEEE, 2001, pp. 29--538.

\end{thebibliography}

\newpage

 





\end{document}